\documentclass{article}

\usepackage{microtype}
\usepackage{graphicx}
\usepackage{subfigure}
\usepackage{booktabs} % for professional tables

\usepackage[table,xcdraw]{xcolor}

\usepackage[textsize=tiny]{todonotes} % Enable todos

\usepackage[english]{babel}

\usepackage{amsmath}
\usepackage{amssymb}
\usepackage{mathtools}
\usepackage{amsthm}

\theoremstyle{plain}
\newtheorem{theorem}{Theorem}

\theoremstyle{definition}
\newtheorem{definition}[theorem]{Definition}

\theoremstyle{remark}

\usepackage{hyperref}

\usepackage{bbm} % For indicator function and blackboard bold symbols

\usepackage{algorithm}
\usepackage{algpseudocode}

\usepackage[capitalize,noabbrev]{cleveref}

\usepackage[accepted]{icml2024} % Use for camera-ready submission

\icmltitlerunning{CausalSent}

\begin{document}

\twocolumn[

\icmltitle{CausalSent\\ Interpretable Sentiment Classification with RieszNet}



\icmlsetsymbol{equal}{*}

\begin{icmlauthorlist}
\icmlauthor{Daniel Frees}{equal,StanfordStats,Google}
\icmlauthor{Martin Pollack}{StanfordStats}
\end{icmlauthorlist}

\icmlaffiliation{StanfordStats}{Department of Statistics, Stanford University}
\icmlaffiliation{Google}{Google, San Francisco, CA}

\icmlcorrespondingauthor{Daniel Frees}{dfrees@stanford.edu}

\icmlkeywords{Machine Learning, ICML}

\vskip 0.3in
]



\printAffiliationsAndNotice{}  % leave blank if no need to mention equal contribution

\begin{abstract}
Despite the overwhelming performance improvements offered by recent natural language processing (NLP) models, the decisions made by these models are largely a black box. Towards closing this gap, the field of causal NLP combines causal inference literature with modern NLP models to elucidate causal effects of text features. We replicate and extend \cite{bansal2023rieszcontrolling}'s work on regularizing text classifiers to adhere to estimated effects, focusing instead on model interpretability. Specifically, we focus on developing a two-headed RieszNet-based neural network architecture which achieves better treatment effect estimation accuracy. Our framework, CausalSent, accurately predicts treatment effects in semi-synthetic IMDB movie reviews, reducing MAE of effect estimates by $2$-$3$x compared to \cite{bansal2023rieszcontrolling}'s MAE on synthetic Civil Comments data. With an ensemble of validated models, we perform an observational case study on the causal effect of the word "love" in IMDB movie reviews, finding that the presence of the word "love" causes a $+2.9\%$ increase in the probability of a positive sentiment.

%In this work, we motivate a subfield of causal NLP, causal regularization, in which our task is to simultaneously learn a predictive NLP model and estimate causal effects of the model's input features. With these learned causal effects, one can constrain the predictive model to adhere to causality and reduce the deleterious effects of spurious correlations in the data. 
    
\end{abstract}

\section{Introduction}

%In the present work we focus on developing an NLP model which simultaneously learns to perform text classification and learns accurate causal effects of text features on the classification outcome. To motivate the relevance of this setting and the design choices in our model, we first present an overview of causal NLP, transfer learning, sentiment analysis (our task), and causal regularization.

\subsection{Causal NLP}

Recent advances in the field of natural language processing (NLP), including the success of transformer-based large language models (LLMs), have yielded massive improvements across classical language tasks such as sentiment analysis, text classification, and more. Despite these advances, the growing complexity of natural language models makes model interpretation extremely difficult, ultimately eroding model trust \cite{ribeiro2016why}. The relatively infant field of causal NLP seeks to address this gap by combining causal inference literature with advanced NLP models \cite{feder2021causalNLP}. Some common use cases include debiasing and regularizing large language models \cite{bansal2023rieszcontrolling, bansal2022usinginterventionsimproveoutofdistribution, veitch2021counterfactual}, interpreting the impact of text features \cite{ravfogel2021counterfactual, sridhar2019estimating, ribeiro2016why}, and performing what-if analyses on synthetic data \cite{veitch2021counterfactual}. To further motivate the task of causal inference in NLP, we provide two specific example settings in \autoref{sec:add-causal-nlp-ex}.

\subsection{Spurious Correlations and Causal Regularization}

Spurious (non-causal) correlations occur frequently in text data, arising from text phrase multicollinearities, sampling biases, random chance and more \cite{gururangan2018annotation}. If NLP models learn to over-adhere to spurious features, their performance will drop significantly on "out-of-domain"/ "out-of-distribution" (OOD) data, where systematic shifts in the underlying data distribution occur. Understanding and enforcing \textit{causal} relationships as opposed to \textit{correlations} can improve both the interpretability of models and their robustness to perform well on new data \cite{feder2021causalNLP, veitch2021counterfactual, bansal2023rieszcontrolling, bansal2022usinginterventionsimproveoutofdistribution}. Causal effect regularization is the task of\textit{ enforcing adherance to causal relationships}. 

To further motivate causal effect regularization, consider the following example: In the IMDB movie review dataset \cite{maas2011imdb}, the presence of the word "love" is correlated with an $+18.4\%$ higher chance of a positive sentiment. In \autoref{tab:SpurCorr} we see a "majority" text example (where the positive correlation holds), but also a "minority" text example (where the correlation breaks down). If a sentiment classification model over-adheres to the correlation between "love" and sentiment, it is likely to predict the second example incorrectly. 

\begin{table*}[!htb]
\centering
\begin{tabular}{|l|l|}
\hline
\rowcolor{gray!20} \textbf{Majority Text} & \textbf{Sentiment Label} \\
\hline
... at the Toronto International Film Festival. I \textcolor{rgb:red,0.1;green,0.8;blue,0.1}{loved} this, and not just for the obvious ... & \textcolor{rgb:red,0.1;green,0.8;blue,0.1}{POSITIVE} \\
\hline
\rowcolor{black!10} \textbf{Minority Text} & \textbf{Sentiment Label} \\
\hline
... he's happily surprised by the girl he \textcolor{blue}{loves}, he's an \textcolor{red}{awful actor} ... & \textcolor{red}{NEGATIVE} \\
\hline
\end{tabular}
\caption{Spurious Correlations in IMDB Movie Review Examples. Green indicates positive use of the word love, blue indicates neutral use of the word love, and red indicates negative predictors of the sentiment label.}
\label{tab:SpurCorr}
\end{table*}

%\cite{ribeiro2016why} motivates the importance of understanding the decision-making processes of machine learning models, citing positive implications for both user trust and downstream model improvement. Towards achieving model interpretability, 

%In a causal NLP framework, we consider that texts can be treatments, confounders, and/or outcomes, creating a wide variety of possible use cases.

\subsection{Sentiment Analysis}
\label{sec:sentiment-analysis}

As a case study, we focus on sentiment analysis, the task of mining emotions, attitudes, and opinions from text. Transformer-based models like BERT \cite{devlin2019bert}, RoBERTa \cite{liu2019roberta}, and DistilBERT \cite{sanh2019distilbert} excel at sentiment analysis, as do LLMs (either few-shot pre-trained or fine-tuned). All of these models are highly complex, making model interpretability difficult. Given its simple outcome variable (sentiment) and complex model choices (LLMs), sentiment analysis serves as an ideal case study for NLP interpretability. 

%Given that transformer-based models like BERT \cite{devlin2019bert}, RoBERTa \cite{liu2019roberta}, and DistilBERT \cite{sanh2019distilbert} excel at sentiment analysis, as do LLMs (either few-shot pre-trained or fine-tuned).

%In early development, researchers used lexicon-based scoring approaches to aggregate word sentiment scores from pre-defined static dictionaries. These were not particularly successful, and recent machine learning and deep learning approaches have completely dominated early methods. In particular, transformer-based models like BERT \cite{devlin2019bert}, RoBERTa \cite{liu2019roberta}, and DistilBERT \cite{sanh2019distilbert} excel at sentiment analysis, as do LLMs (either few-shot pre-trained or fine-tuned). While LLMs perform well on simple sentiment scoring tasks, they struggle on more complex tasks \cite{zhang2023sentiment} and failure points are difficult to identify given their complexity. Sentiment analysis lends itself as a natural task for evaluating causal regularization methods, as the sentiment outcome variable is simple and easy to understand, so research can focus on the complexity of the textual treatments and confounders. 

%In particular, causal methods might be promising for increasing performance on texts where common correlations no longer hold true (e.g. irony), as well as increasing overall model interpretability. 

\subsection{Text Matching}

Though we focus here on sentiment analysis, causal effect regularization techniques (and related heuristic techniques) have also been deployed successfully to combat the effects of spurious correlations in other fields of NLP, including text matching. See \autoref{ref:sec:regularize-embed} for details.

\subsection{Unique Challenges in Causal NLP}

In addition to frequent spurious correlations, natural language data presents numerous unique challenges that make causal inference hard. There is no clear definition for a treatment: is treatment the presence or absence of a phrase, is it phrase $A$ versus phrase $B$, is it the style or some other attributed extracted from the input text? In causal inference, one can typically cleanly separate the treatment variable $T$ from the other variables, but that is not the case in NLP, as words and phrases mix with their entire context to produce embeddings. Text data is also complex, high-dimensional, and sparse. This yields overlap issues \cite{damour2020overlap} and increases the computational and mathematical complexity of models. The massive vocabulary size of any given language also means that models are unlikely to be trained against a comprehensive sample of language, and OOD shifts become likely. All of these challenges are exarcerbated by lengthier texts.

%The unique challenges of textual data and the field of NLP present challenges that must be considered or addressed. To begin, even the basic definition of causality is ambiguous when applied to textual data. We do not usually think of individual words and phrases directly affecting an outcome, but rather whole texts tend to produce meaning. Similarly, how should we define a treatment? In this approach, we proceed with the following definitions, which can still be debated: we take causality to mean how the choice of words effects a certain outcome like sentiment or the probability of two texts matching, and then treatment is defined as the inclusion or absence of a selected treatment phrase.

% An additional challenge stems from the fact that textual data is extremely complex, high-dimensional, and sparse. The vocabulary of a single language is massive, and our models need to be able to handle the inclusion of any word. Also, any given combination of words is highly specific, and a certain idea can be captured with text in any number of ways. This means that it is unlikely to see certain words together in a sentence in multiple ways, leading to a major issue when we consider the inclusion of co-variates, or words in our texts that are not our treatment phrase: the assumption of overlap \cite{damour2020overlap}. We will most likely not have examples in our dataset of two identical pieces of text, with one having the treatment phrase and the other not. Thus, popular causal inference techniques which rely on this assumption are not valid for our use case.

\section{Related Works}

\subsection{General Deep Learning for Causal Inference}

In 2019, \cite{shi2019dragonnet} proposed DragonNet, a deep learning framework for estimating causal effects from observational data. \cite{shi2019dragonnet} uses neural networks (with some extra optimizations) to estimate expected outcomes and propensity scores. They combine these outputs to form a doubly robust propensity-weighted estimator for the ATE. 

%An impactful work on the use of neural networks to learn causal effects for generic data, DragonNet  builds a propensity-based estimator for the average treatment effect (ATE) of an input feature. However, this method was not applied to textual data which has low overlap, meaning propensity scores approach $1$ or $0$. In this scenario, propensity-based estimators suffer from instability and high variance as the estimate contains propensity scores in the denominator.

Unfortunately, because propensities ($Pr[T|X]$) approach $0$ or $1$ in high-dimensional text data, propensity based estimators suffer from instability and high variance (see \autoref{par:propensity}). To combat the instability issues of propensity-based techniques, one can consider directly learning Riesz representers instead. Improving upon DragonNet, \cite{chernozhukov2021riesznet} proposes RieszNet, a neural architecture which learns expected outcomes and Riesz representers simultaneously. These outputs can then be combined to form a doubly robust Riesz estimator for the ATE. 

%Several recent papers have sought to improve machine learning (ML) systems' performance on out-of-distribution data, where the underlying distribution of inputs or outputs shifts significantly. These works include general causal ML frameworks for deep learning of causal effects, efforts to improve text feature extraction by removing spurious information and applied causal ML and algorithmic frameworks for language data.

\subsection{Causal Effect Regularization and Interpretability}

Causal effect regularization and model interpretability can be achieved via both interventional approaches (interventional regularization) and causal-inference based approaches (causal regularization). 

\paragraph{Interventional Regularization}
\cite{veitch2021counterfactual} investigates causal regularization techniques in the general ML setting, motivating "counterfactual invariance" around irrelevant input dimensions as a signal of a well-regularized model that will be more robust for OOD model performance. To measure counterfactual invariance \cite{veitch2021counterfactual} performs 'stress test' perturbations by producing counterfactual model inputs and measuring resulting differences in outcomes. Informed by knowledge about which variables should and should not affect the outcome, stress tests can be employed to ensure that unimportant variables minimally affect outputs. In NLP, interventional techniques include embedding dropout and token masking. To measure the causal effect of text features on predictions, words are masked or latent features are dropped from the embedding, and the causal effects are defined to be the resulting change in the outcome variable. \cite{ravfogel2021counterfactual} estimates causal effects of text features via dropout techniques with BERT embeddings, demonstrating that these effects can reveal information about model logic.  \cite{bansal2022usinginterventionsimproveoutofdistribution} estimates causal effects of phrases on text semantics using masking techniques and regularizes phrase effects during model training to bolster fine-tuned embedding models against OOD shifts.

\paragraph{Causal Regularization} Causal regularization can similarly be achieved via a two-step process. First, estimate accurate feature effects. Second, introduce a penalty or some other constraint into model tuning to enforce estimated feature effects. Whereas interventional techniques perform perturbations (interventions) to measure effects, causal techniques employ estimators from the field of causal inference. 

To estimate the causal effect of words on sentiment and toxicity \cite{bansal2023rieszcontrolling} proposes a RieszNet-style \cite{chernozhukov2021riesznet} deep learning model which simultaneously predicts classification outcomes and Riesz representers. These model outputs can be combined to estimate doubly robust causal effects, and the classification model can then be regularized by either incorporating an MSE loss term that penalizes deviation from the estimated effects, or by mixing synthetic counterfactuals into the training dataset. \cite{bansal2023rieszcontrolling} focus primarily on regularization via synthetic counterfactuals, denoting this process of estimating feature effects and augmenting the training data \texttt{FEAG}. 

\cite{bansal2023rieszcontrolling} evaluate their \texttt{FEAG} pipeline for two sentiment analysis datasets: CivilComments toxicity detection \cite{borkan2019civilcomments}  and IMDB movie review sentiment \cite{maas2011imdb}. They compare their results with popular baselines such as \citep{joshi2022spurious, he2022bias, orgad2022debiasing}, weighting methods like DFL, DFL-nodemog, Product of Experts \citep{mahabadi2019bias, orgad2022debiasing}, and latent space removal methods like INLP \citep{ravfogel2020null}. Their novel causal regularization method performs on par with the aforementioned baselines for both IMDB sentiment classification and Civil Comments toxicity detection. However, when considering specific partitions of the data, \texttt{FEAG} improves the within-group accuracies of the following sets of observations by up to 10\%: $(Y=1, T=1), (Y=0, T=0), (Y=1, T=0), (Y=0, T=1)$. The authors propose that this within-group improvement is the result of better prediction performance on minority texts (such as the second row of \autoref{tab:SpurCorr}). 

\paragraph{Interpretable NLP}
Other works focus primarily on the first step of causal regularization, which might be appropriately named \textit{model interpretability}. Towards estimating average treatment effects (ATEs) in observational data with textual confounders, \cite{roberts2020adjusting} proposes a matched pairs estimator, denoted topical inverse regression matching, which learns a low-dimensional representation containing propensity and topical text information to support subsequent text matching. Similarly, \cite{veitch2020adapting} explores supervised dimensionality reduction and language modeling techniques to extract "causally sufficient" embeddings (embeddings which are predictive of both outcome and treatment) for confounder adjustment, finding that both approaches improve causal effect estimation performance on semi-synthetic datasets. Whereas \cite{roberts2020adjusting} proposes a matched pairs estimator, \cite{veitch2020adapting}'s approach is appropriate for doubly robust estimators using propensity scores or Riesz representers. In contrast to the previous works which estimate the ATE, \cite{ribeiro2016why} proposes LIME, a submodular optimization algorithm for deriving "locally interpretable" model approximations to better understand model decisions. Lastly, \cite{sridhar2019estimating} demonstrate that text data can be used simultaneously as \textit{treatment, confounder, and effect}. Specifically, \cite{sridhar2019estimating} use various feature extraction techniques to show that tone in online debates is predictive of subsequent dialogue after adjusting for participant ideologies modeled from text data.

\subsection{Literature Gap}

We aim to replicate and extend the work of \cite{bansal2023rieszcontrolling}. Whereas \cite{bansal2023rieszcontrolling} focuses primarily on improving text classification model robustness via causal regularization, we focus on interpretable NLP and improved causal effect estimation accuracy. Specifically, we implement a wide variety of new architecture options and run numerous grid search experiments to determine an optimal architecture, learning regime, and other hyperparameters for accurate ATE estimation. 

We note several limitations in \cite{bansal2023rieszcontrolling}. First, no semi-synthetic validation experiments were reported for IMDB, and no semi-synthetic validation experiments were reported for negative target effects\footnote{In early testing, we found that some models are only able to accurately estimate either positive or negative effects, so symmetric validation is important}. We focus our work on the IMDB data, and include a symmetric range of target effects.  Second, their synthetic Civil Comments data is generated with artificially inflated overlap, and the synthetic data relies on the assumption that treatment text $T$ and covariate text $X$ are correlated only through author intent $C$, which we find questionable. We remove this assumption in our causal graph, and empirically validate treatment effect estimations in the naturally occurring low-to-no overlap setting. Despite violating core assumptions (\autoref{sec:assumptions}), we feel this setting is more realistic in actual application. Furthermore, while \cite{bansal2023rieszcontrolling} achieve fairly good feature effect estimates in high overlap synthetic data, their performance decreases dramatically in the $1\%$ overlap setting, with mean absolute errors ranging to up $59\%$ relative to the true target effect. We seek to improve upon this result. 

Finally, despite producing interesting results, \cite{bansal2023rieszcontrolling} do not neatly package or share an open-source implementation of their model. We publish our implementation at \url{https://github.com/danielfrees/causalsent}.

\section{Data}

\paragraph{IMDB Movie Reviews}

\begin{figure}[h!]
    \centering
    \includegraphics[width=0.7\linewidth]{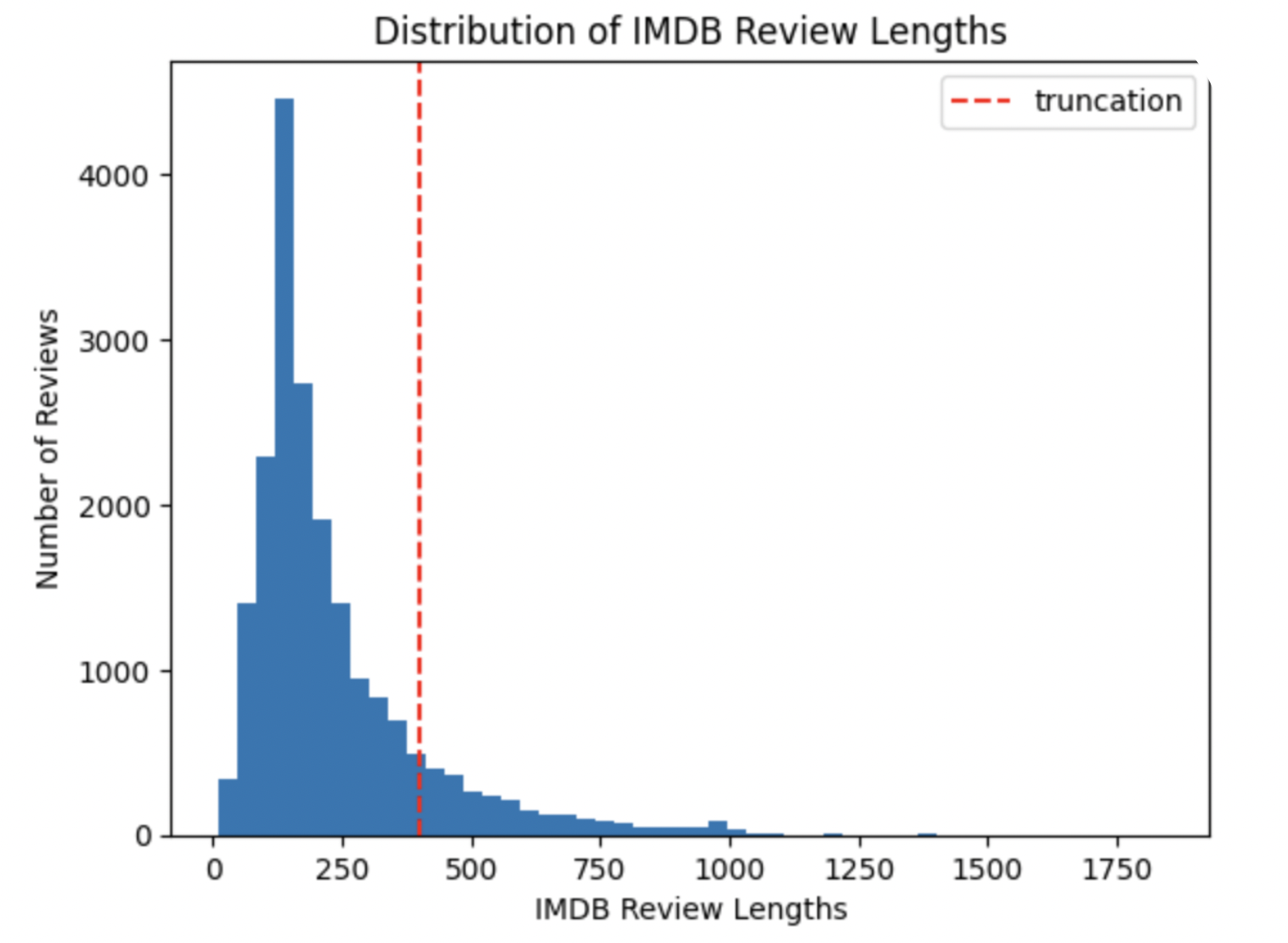}
    \caption{Distribution of IMDB Review Lengths}
    \label{fig:imdb_review_lengths}
\end{figure}

The IMDB Movie Review data \cite{maas2011imdb} consists of $50,000$ highly polarized movie reviews with binary sentiment labels (Pos: $1$, Neg: $0$). To preprocess the data, we tokenize the texts, heuristically select a treatment phrase, synthesize counterfactual inputs as described in \autoref{sec:gen-counterfactual-inputs}, optionally manipulate the true average treatment effect as described in \autoref{sec:synth-ovr}, and add a custom data collation for batched model training. We randomly split the original $25,000$ train examples into $80\%$ training data ($20,000$ examples) and $20\%$ validation data ($5,000$ examples). We reserve the original $25,000$ test examples for sentiment task evaluation. We truncate texts to $150$ tokens during synthetic experiments due to computational limits, and truncate to $400$ tokens on our final observational model experiments. The latter approximate truncation (assuming 1:1 word to token conversion) is visualized against the distribution of text lengths in \autoref{fig:imdb_review_lengths}.

\paragraph{CivilComments Toxicity Texts}

The Civil Comments dataset \cite{borkan2019civilcomments} consists of $1.8M$ comments from the Civil Comments platform, a commenting plugin for independent English-language news sites. Labels are real-valued toxicity scores ranging from $0$ to $1$. Labels are heavily skewed towards $0$, with $540,110$ non-zero toxicities. As such we limit the dataset to $540,110$ 'toxic' non-zero toxicities, binarized to a label of $1$, and $540,110$ 'non-toxic' comments randomly sampled from the $0$s. For our Civil Comments experiments, we further limit the data to $30,000$ examples in each category due to computational limits. We truncate lengths to $150$ tokens.

\begin{figure}[h!]
    \centering
    \includegraphics[width=0.7\linewidth]{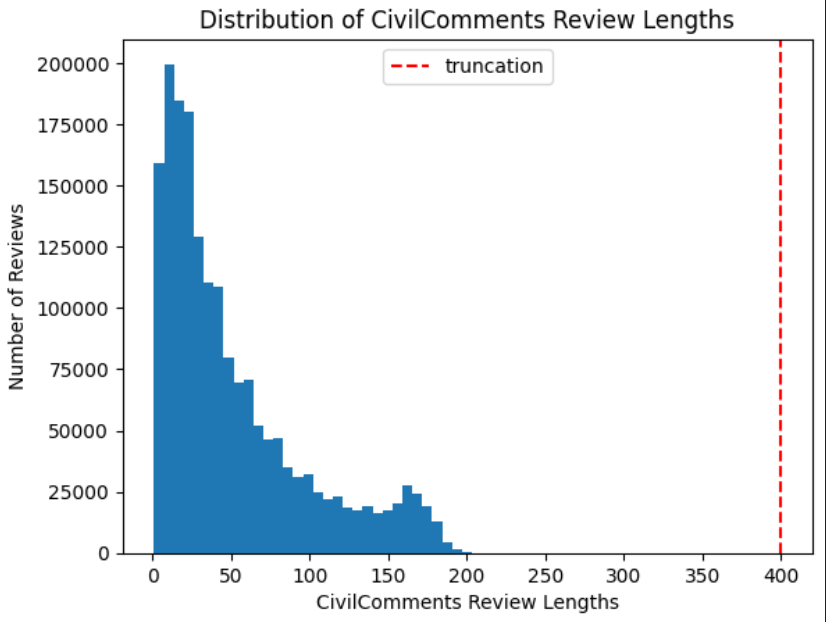}
    \caption{Distribution of Civil Comments Text Lengths}
    \label{fig:civil_review_lengths}
\end{figure}

\subsection{Generating Counterfactual Inputs}
\label{sec:gen-counterfactual-inputs}

We consider the simplest definition of treatment in text: presence of a treatment phrase. As such, to generate positive counterfactual inputs $(X, 1)$, we prepend treatment word $T$ to the text $Z$ if it is not present. To generative a negative counterfactual $(X, 0)$, we mask the treatment word $T$ anywhere it appears in the text $Z$.

\subsection{Synthetic Datasets}
\label{sec:synth-ovr}

To induce synthetic true treatment effects, we prepend an artificial treatment word such as "artichoke" to a portion of our texts, and flip outcome labels via a Bernoulli process to induce an artificial treatment effect for the word "artichoke". This approach is similar to \cite{bansal2023rieszcontrolling}'s Semi-Synthetic Civil Comments approach, except we follow our above definition of counterfactuals (\autoref{sec:gen-counterfactual-inputs}) instead of using antonyms “Treated” vs. “Untreated”, and we do not artificially modify overlap by modeling author intent, as \cite{bansal2023rieszcontrolling} does. Despite potentially violating causal modeling assumptions, we feel that our setting is more reflective of real-word data as we do not interfere with the underlying distribution of confounding texts. Despite modifying the underlying data distribution less than \cite{bansal2023rieszcontrolling}, we still feel that prepending treatment words is a very artificial data setting, as we mention in our next steps and limitations (\autoref{sec:limitations}). For details on our synthetic dataset algorithm, see \autoref{sec:synthetic-alg}.

\section{Methods}

\subsection{Setup}

\begin{figure}[!h]
    \centering
    \includegraphics[width=0.9\linewidth]{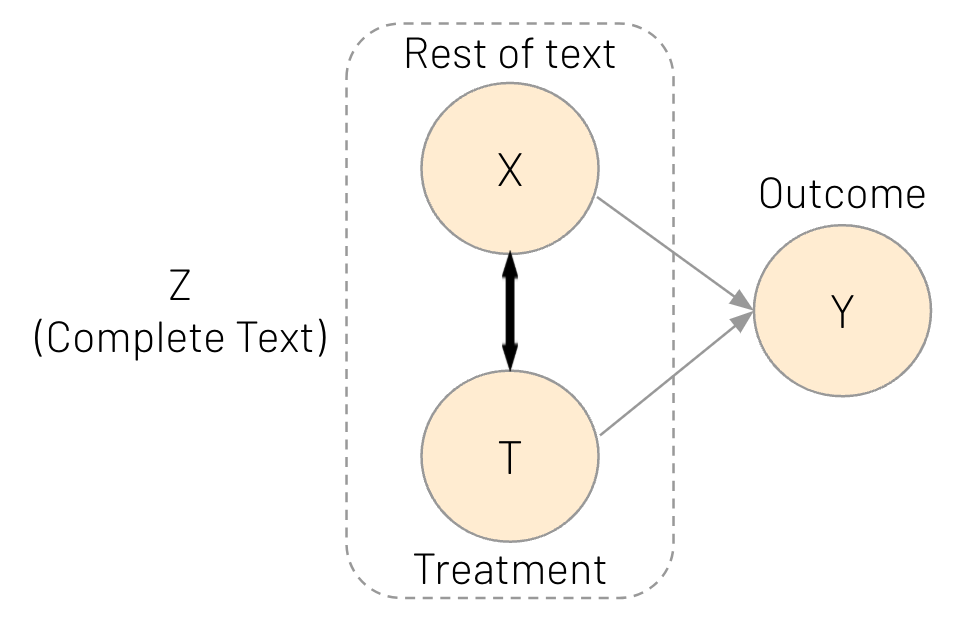}
    \caption{Our causal graph. Adopted from Bansal et. al 2023. We drop the assumption that $X$ and $T$ are only correlated through the author intent $C$.}
    \label{fig:causalgraph}
\end{figure}

To describe the causal process for generating a text with a particular sentiment, we propose a causal graph similar to \cite{bansal2023rieszcontrolling}, pictured in \autoref{fig:causalgraph}. The movie review text $Z$ can be separated into a heuristically chosen treatment phrase, $T$, and all other words in the text, $X$. Both $T$ and $X$ affect the sentiment, $Y$. The backdoor path between \( T \) and \( Y \), \( T \leftrightarrow X \rightarrow Y \), can be blocked by conditioning on \( X \), allowing identification of the causal effect of \( T \) on \( Y \).

\begin{figure}[!h]
    \centering
    \includegraphics[width=0.6\linewidth]{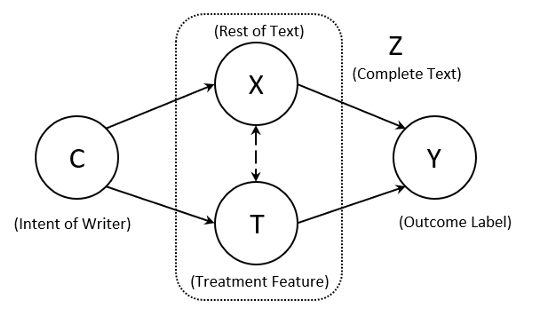}
    \caption{Bansal et. al 2023's causal graph, which assumes that $X$ and $T$ are only correlated through the author intent $C$.}
    \label{fig:causalgraph-bansal}
\end{figure}

There may also be unobserved variable $C$, the intent of the writer, which affects the movie review sentiment $Y$, but only through the text of the movie review $Z$, as there is no direct interaction between writer and reviewer. In this setup, conditioning on $X$ still suffices to identify the effect of $T$ on $Y$. The assumption by \cite{bansal2023rieszcontrolling} that confounding between $X$ and $T$ is mediated entirely by $C$ (\autoref{fig:causalgraph-bansal}) seems questionable, as language is well-known to be autoregressive. Thus, we do not replicate the portion of \cite{bansal2023rieszcontrolling}'s work which aims to model author intent $C$, and instead focus on controlling for confounding language $X$.

\subsection{Treatment Effect Estimation}

We aim to estimate the average treatment effect (ATE) of our treatment text feature(s), e.g. 'love', on text classification, e.g. sentiment. A naive approach to learning this treatment effect would be to train a model, $g$, to approximate the true/oracle data distributing function using our true labels $Y$

\begin{equation}
    g = \arg\min_g \mathbb{E}_\mathcal{D}[\mathcal{L}(Y, g(Z))].
\end{equation}

and directly estimate the average treatment effect:

\begin{equation}
\widehat{\text{ATE}}_{\text{Direct}} = \frac{1}{n} \sum_{i} \left( g(X_i, 1) - g(X_i, 0) \right)
\end{equation}

However, our treatment variable $T$ is correlated to the text covariates $X$, creating bias in the direct model ATE. The treatment $T=$ "love", for example, is much more likely to co-occur with the phrases "romance" and "flowers", than with a description of a knee surgery. As mentioned in our setup, $T$ and $Y$ are $d$-separated by $X$ so by conditioning on $X$, we can achieve identification of the effect of $T$ on $Y$. 

\begin{theorem}[Sufficiency of Propensity Score]
\label{thm:prop}
If the average treatment effect $\psi$ is identifiable from observational data by adjusting for $X$, i.e.,
\[
\psi = \mathbb{E}[\mathbb{E}[Y \mid X, T = 1] - \mathbb{E}[Y \mid X, T = 0]],
\]
then adjusting for the propensity score $e(X)$ also suffices:
\[
\psi = \mathbb{E}[\mathbb{E}[Y \mid e(X), T = 1] - \mathbb{E}[Y \mid e(X), T = 0]].
\]
\end{theorem}

Given \autoref{thm:prop}, it also suffices to condition on the propensity score $e(x) = Pr[T|X]$. This motivates a different, now unbiased ATE where we weight each outcome by the propensity of treatment, such that treated outcomes are upweighted where propensity is low, and vice versa. 

\begin{equation}
\widehat{\text{ATE}}_{\text{propensity}} = \frac{1}{n} \sum_{i} \alpha_{e}(Z_i) Y_i.
\end{equation}

\noindent where
\begin{equation}
\alpha_{e}(Z_i) = \left( \frac{T_i}{e(X_i)} - \frac{1 - T_i}{1 - e(X_i)} \right).
\end{equation}

\noindent defining propensity as 
\begin{equation}
e(X_i) = Pr[T|X_i]
\end{equation}

\label{par:propensity}
Even better, we could form a doubly robust propensity estimator such as AIPW \cite{robins1994estimation}. Unfortunately, the propensity weighting approach will yield extremely high variance in the long-form text data setting, since overlap is likely to be extremely low, yielding propensities close to $0$ or $1$ and thus exploding the propensity weights. Directly learning Riesz Representer weights is therefore a much better strategy as we can learn analogous weights $\alpha_{R}(.)$ directly rather than by estimating propensities as an intermediate step and plugging into a potentially unstable weight equation \cite{chernozhukov2021riesznet}. 

But the question still remains: how do we directly estimate $\alpha_R(.)$? We can use the following Riesz Representer Theorem from \cite{chernozhukov2018automatic} to write our desired average treatment effect in a new manner:
\begin{theorem}
\label{thm:riesz}
(Riesz Representer Theorem). For a square integrable function $f(Z)$ (i.e. $E[f^2(Z)] <\infty$), there exists a square integrable function $\alpha_R(Z)$ such that

$$E[m((Y, Z); f)] = E[\alpha_R(Z)f(Z)]$$

if and only if $E[m((Y, Z); f)]$ is a continuous linear functional of $f$.
\end{theorem}

Applying \autoref{thm:riesz} and defining the ATE as $\theta$, we can write
$$
\theta = E[g(X, 1) - g(X,0)] = E[\alpha_R(Z) g(Z)].
$$

As is done with RieszNet (\cite{chernozhukov2021riesznet}), we learn a neural network head, $\hat{\alpha_R}(Z)$, which outputs a value for the Riesz Representer given an input text $Z$. However, it is still unclear what loss function we should use to fit a Riesz Representer. Following the derivation in \cite{chernozhukov2021riesznet}, we can learn $\hat\alpha_R$ using the optimization below which encourages $\alpha$ to approach the true $\alpha_R$ via an MSE loss:

\begin{align*}
\hat{\alpha}_R &= \arg\min_{\alpha} \mathbb{E}\left[(\alpha_R(Z) - \alpha(Z))^2\right] \\
&= \arg\min_{\alpha} \mathbb{E}\left[\alpha_R(Z)^2 - 2\alpha_R(Z)\alpha(Z) + \alpha(Z)^2\right] \\
&= \arg\min_{\alpha} \mathbb{E}\left[-2\alpha_R(Z)\alpha(Z) + \alpha(Z)^2\right] \\
&= \arg\min_{\alpha} \mathbb{E}\left[-2(\alpha(X, 1) - \alpha(X, 0)) + \alpha(Z)^2\right]
\end{align*}

On the second line, we expand the quadratic. On the third, we remove the ground truth $\alpha_R$ since we are not minimizing over that term. Lastly, we apply the Riesz Representer Theorem using $\alpha$ in place of $f$ in the definition. This results in an optimization only in terms of $\alpha$ and counterfactual input data.

Finally, we can form an unbiased doubly robust estimator by combining our learned classification regression $g(\cdot)$ and Riesz representer $\hat{\alpha}_R(\cdot)$ models, and using the typical doubly robust approach:
\begin{align*}
\widehat{\text{ATE}}_{\text{DR,R}} &= \widehat{\text{ATE}}_{\text{Direct}} + \frac{1}{n} \sum_{i=1}^n \hat\alpha_R(Z_i)(Y_i - g(Z_i))
\end{align*}

\subsection{Model Optimization}

Now we have derived the estimator for the ATE of our selected treatment phrase. We showed that the loss function for fitting the Riesz head of our neural network is
\begin{align*}
L_{\text{Riesz}} &= \mathbb{E}[-2(\alpha(X, 1) - \alpha(X, 0)) + \alpha(Z)^2]
\end{align*}

We also need to learn our sentiment predictor $g(\cdot)$. With a binary sentiment outcome, we use a binary cross entropy loss to form our "sentiment loss":
\begin{align*}
    L_{\text{Sentiment}} &= \mathbb{E}[-Y \log(g(Z)) - (1 - Y) \log(1 - g(Z))]
\end{align*}

We then add an L1 regularization, as suggested by \cite{chernozhukov2018automatic}, to all trainable parameters. If we let $\Theta_g$ be the parameters in the sentiment head, $\Theta_\alpha$ the parameters in the Riesz Head, and $\Theta_\beta$ the unfrozen parameters in our LLM feature extractor backbone, this yields an "L1 loss" of the form
\begin{align*}
    L_{\text{L1}} &= \sum_{\theta_g \in \Theta_g} |\theta_g| + \sum_{\theta_\alpha \in \Theta_\alpha} |\theta_\alpha| + \sum_{\theta_\beta \in \Theta_\beta} |\theta_\beta|
\end{align*}

Lastly, our main objective in causal regularization is to constrain the output of our $g(.)$ function to maintain causal relationships. We use an MSE loss to constrain $g$'s estimated difference in counterfactuals to be close to our estimated causal effect. 

\begin{align*}
    L_{\text{REG}} &= \mathbb{E}\left[(g(X, 1) - g(X, 0) - \hat\tau)^2\right]
\end{align*}

Finally, we combine these four loss functions into an overall loss with the use of four weights. 

\begin{align*}
    L_{\text{Overall}} &= \lambda_{\text{BCE}} L_{\text{Sentiment}} + \lambda_\alpha L_{\text{Riesz}} + \lambda_{\text{L1}} L_{\text{L1}} + \lambda_{\text{REG}} L_{\text{MSE}},
\end{align*}

and our estimated functions can be found with 
\begin{align*}
    \hat{g}, \hat{\alpha}, \hat{\beta} &= \operatorname*{arg\,min}_{g, \alpha, \beta} L_{\text{Overall}}
\end{align*}

% \subsection*{Regularization: Augmentation}
% Given inputs $Z = (X, T)$, we build counterfactual inputs $Z^c = (X, 1 - T)$. \\
% We then perform a label flipping algorithm that takes into account $\hat{\tau}$ to produce counterfactual outputs $Y^c$. This yields a counterfactual dataset $\mathcal{D}^c$ to complement our original dataset $\mathcal{D}$.\\
% \begin{align*}
% \operatorname*{arg\,min}_f \mathbb{E}_\mathcal{D}[\mathcal{L}(Y, f(Z))] + \lambda \mathbb{E}_{\mathcal{D}^c}[\mathcal{L}(Y, f(Z))]
% \end{align*}

\subsection{Model Architecture}

As demonstrated by both \cite{shi2019dragonnet} and \cite{chernozhukov2021riesznet}, an effective architecture for deep learning of causal effects consists of a single pre-trained backbone, concatenated with two different final layer linear heads, one to learn the Riesz Representer and another to learn the downstream task (e.g. sentiment prediction). We adopt this shared LLM backbone architecture, and feed the pooled LLM embeddings\footnote{or all embeddings in the case of the convolutional Riesz and/or sentiment heads} into a sentiment head and Riesz Representer head for simultaneous learning of classification, $g(\cdot)$, and Riesz Representer function, $\hat\alpha_R(\cdot)$. 

\begin{figure}[H]
    \centering
    \includegraphics[width=0.8\linewidth]{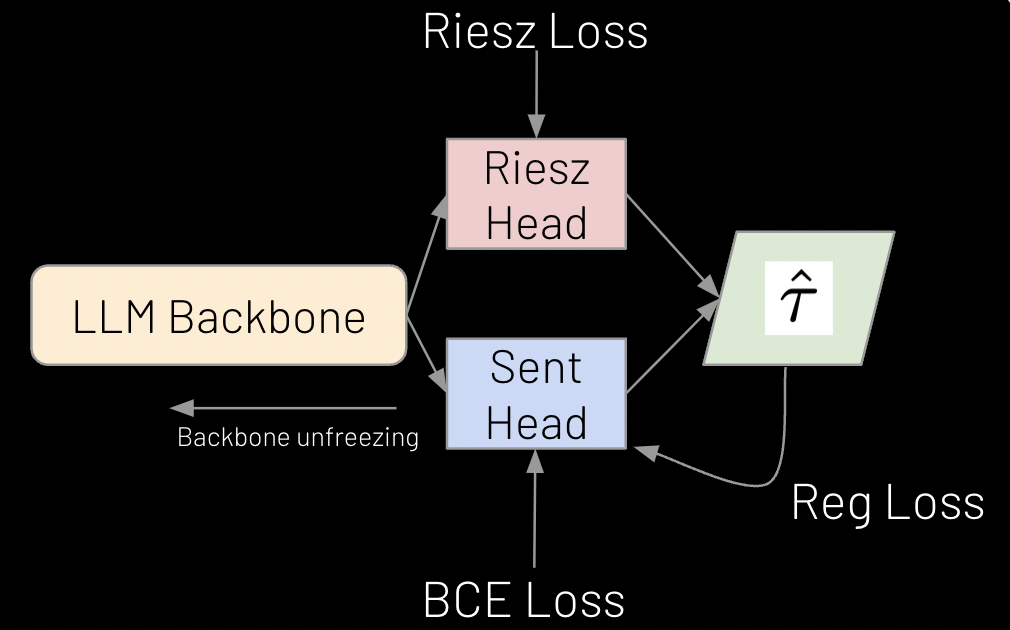}
    \caption{CausalSent Model Architecture}
    \label{fig:CausalSentModelArchitecture}
\end{figure}

Our hope is that the LLM embeddings will be causally sufficient for effect estimation akin to the supervised dimensionality reduction and language modeling techniques employed by \cite{veitch2020adapting}. See \autoref{fig:CausalSentModelArchitecture} for an architecture diagram. To compare, we also created a two-backbone model architecture, where each head is attached to a separate backbone as in \autoref{fig:TwoBackboneCausalSentArchitecture}. For more details on the process of learning with a pre-trained backbone, see \autoref{sec:llm-transfer-learning}.

To extend the work of \cite{bansal2023rieszcontrolling} we introduce a number of novel architecture settings and hyperparameters. We build three different RieszHead and SentimentHead architectures: convolutional, fully-connected deep neural network, and linear. We also implement various backbone unfreezing strategies to enable backpropagation to enable LLM encoder learning. During training we can unfreeze a static \texttt{top\{n\}} layers, or perform iterative unfreezing in the feature extractor, which might be optimal from a performance perspective by mitigating "catastrophic forgetting", improving transfer learning, and protecting against out-of-distribution (OOD) performance shifts \cite{liu2024fun}. We also support dynamic selection of treatment words, loss function weighting to balance between our $4$ losses, direct vs. doubly robust Riesz ATE estimation, DistilBERT \cite{sanh2019distilbert} or llama3 \cite{touvron2024llama3} based backbones, early stopping settings, and various optional settings for synthetic data setup. Lastly, we offer the option to use "interleaved" training, whereby training alternates between the sentiment and Riesz objectives at each epoch.

\subsection*{Two-Backbone CausalSent}

\begin{figure}[H]
    \centering
    \includegraphics[width=0.72\linewidth]{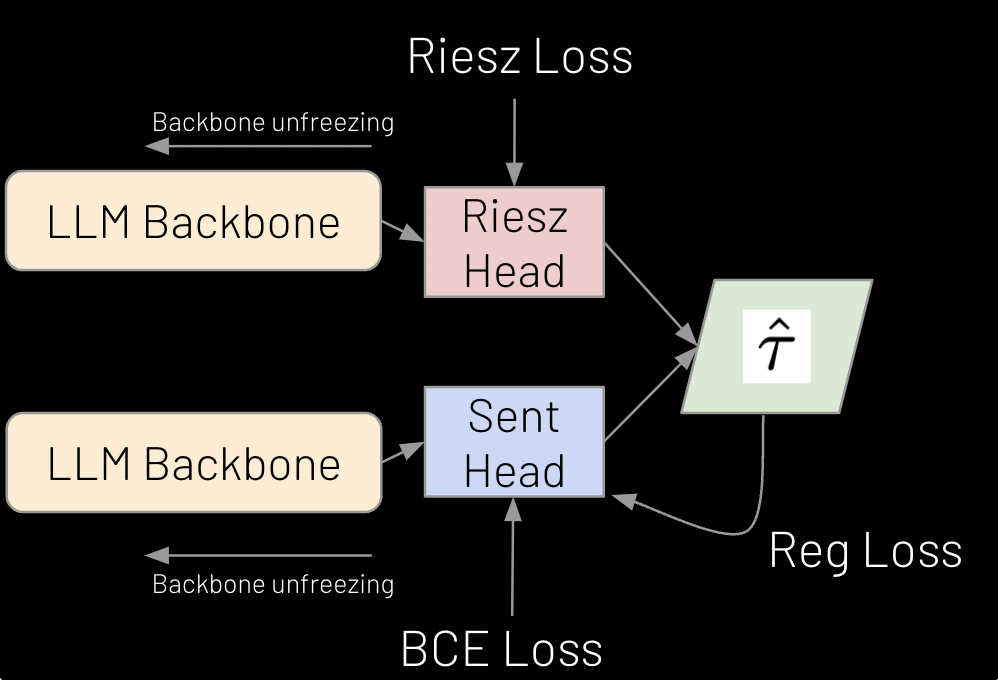}
    \caption{Two-Backbone CausalSent Architecture}
    \label{fig:TwoBackboneCausalSentArchitecture}
\end{figure}

After observing that an unfrozen backbone hugely improved sentiment prediction performance but yielded inaccurate treatment effect estimates, we considered a separate model architecture with separate backbones for the Riesz head and sentiment head (See \autoref{fig:TwoBackboneCausalSentArchitecture}). We froze the Riesz backbone for ATE estimation stability, and unfroze\footnote{partially, or iterative} the sentiment backbone for expressivity. 

\subsection{Hardware}

All experiments were trained using PyTorch \cite{paszke2019pytorch} on a $30$ core Apple Silicon M3 Max GPU.

\section{Results}

Initial experiments for loss weighting and learning rate found that a $\lambda_{\text{BCE}}$ of $0.1$ to $1.0$, $\lambda_{\text{Riesz}}$ loss of $0.1$ to $10.0$, $\lambda_{L1}$ and $\lambda_{REG}$ of $0$ to $0.1$ were optimal based on training stability and sentiment accuracy, alongside a learning rate of 5e-5 and the AdamW optimizer \cite{loshchilov2019decoupled}. Initial architecture experiments found that a fully-connected Riesz head, fully-connected sentiment head, static top $3$ backbone layer unfreezing, treatment word "love", early stopping patience=$4$ and $\epsilon=0.05$ performed best on classification\footnote{Based on recent works which have shown that modern LLM encodings are effective for downstream task learning \cite{bhatia2023tart}, we hypothesize that llama3 might provide superior performance for both effect estimation and sentiment as compared with the 'small' LM DistilBERT. Unfortunately, training our llama3 based models was too slow with available hardware.}, achieving a test accuracy of $0.874$ and test F1 of $0.872$ on IMDB sentiment prediction accuracy. 

\subsection*{Semi-Synthetic ATEs + Simultaneous Learning}

To validate feature effect estimation accuracy, we generate semi-synthetic IMDB datasets with true treatment effects of $-0.5$, $-0.25$, $0.25$, and $0.5$, and simultaneously learn our Riesz Representer and sentiment heads across various hyperparameters and architectures, varied as described in \autoref{tab:synth_sim_gridsearch_experiment}. With the shared backbone model, we achieve accurate treatment effects using a linear sentiment head, but not with a fully-connected (or convolutional) sentiment head, and note small to moderate variance across other hyperparameters. See \autoref{tab:results-1}.

We also test the two-backbone model, finding that it produces inaccurate effect estimates, possibly due to discrepancies in the input features between $\alpha(\cdot)$ and $g(\cdot)$. 

\begin{table*}[!h]
\centering
\begin{tabular}{|c|cc|cc|}
\hline
\rowcolor{gray!20} \textbf{True ATE} & \multicolumn{2}{c|}{\textbf{Estimated Effect, Mean (Min / Max)}} & \textbf{MAE (Linear)} & \textbf{MAE / Target} \\
\cline{2-4}
\rowcolor{gray!20} & \textbf{Linear SentHead} & \textbf{FCN SentHead} & & \\
\hline
-0.5 & \textbf{-0.474} (-0.55 / -0.42) & -0.074 (-0.11 / -0.04) & 0.0547 & 10.94\% \\
-0.25 & \textbf{-0.274} (-0.28 / -0.26) & -0.055 (-0.09 / -0.02) & 0.0243 & 9.72\% \\
0.25 & \textbf{0.271} (0.22 / 0.32) & -0.012 (-0.02 / 0.00) & 0.0491 & 19.64\% \\
0.5 & \textbf{0.465} (0.35 / 0.59) & 0.030 (0.02 / 0.04) & 0.1122 & 22.44\% \\
\hline
\end{tabular}
\caption{Semi-Synthetic ATEs with Simultaneous Learning}
\label{tab:results-1}
\end{table*}

\subsection*{Semi-Synthetic ATEs + Interleaved Learning}

We repeat the previous experiment, but using our "interleaved" training regime, hypothesizing that we might observe more stability and robustness across hyperparameters since we maximize \textit{either} the sentiment or Riesz objectives at each step of stochastic gradient descent. Overall, we do observe less variance in our estimates, but the average causal effect estimates are less accurate. See \autoref{tab:gridsearch_imdb_interleaved} for grid search details and \autoref{tab:synth-interleaved-results} for results.

\begin{table*}[!h]
\centering
\begin{tabular}{|c|cc|cc|}
\hline
\rowcolor{gray!20} \textbf{True ATE} & \multicolumn{2}{c|}{\textbf{Estimated Effect, Mean (Min / Max)}} & \textbf{MAE (Linear)} & \textbf{MAE / Target} \\
\cline{2-4}
\rowcolor{gray!20} & \textbf{Linear SentHead} & \textbf{FCN SentHead} & & \\
\hline
-0.5 & \textbf{-0.387} (-0.40 / -0.38) & 0.099 (0.09 / 0.10) & 0.1127 & 22.54\% \\
-0.25 & \textbf{-0.319} (-0.33 / -0.31) & 0.246 (0.22 / 0.31) & 0.0689 & 27.58\% \\
0.25 & \textbf{0.200} (0.20 / 0.21) & 0.633 (0.53 / 0.91) & 0.0498 & 19.91\% \\
0.5 & 0.375 (0.37 / 0.39) & \textbf{0.512} (0.45 / 0.68) & 0.1254 & 25.09\% \\
\hline
\end{tabular}
\caption{Semi-Synthetic ATEs with Interleaved Learning}
\label{tab:synth-interleaved-results}
\end{table*}

\subsection*{Semi-Synthetic ATEs + Interleaved Learning + Embedding Learning}

\begin{figure}[h!]
\centering
\includegraphics[width=1\linewidth]{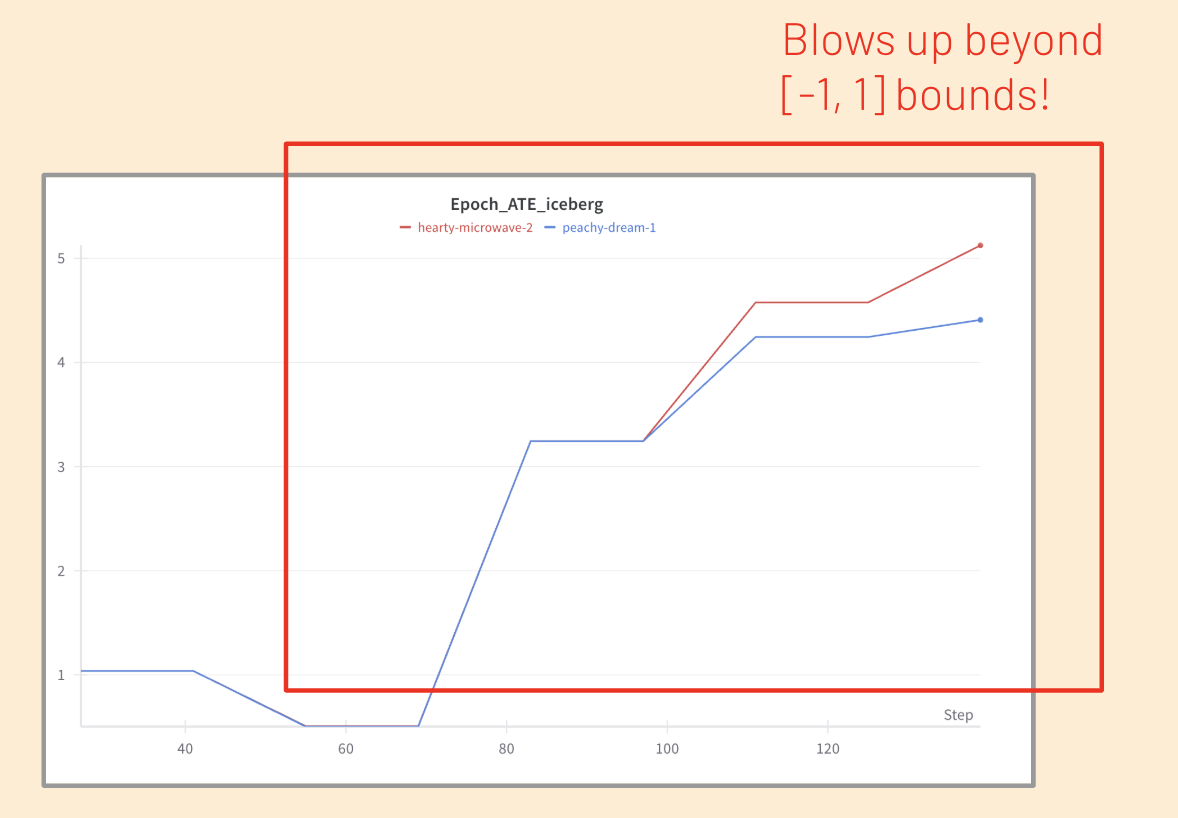}
\caption{Unfreezing the LLM backbone yields instability and inaccurate causal effect estimates. Pictured are the full-epoch ATE estimates throughout interleaved, top$3$ backbone layers unfrozen training.}
\label{fig:unfreeze_blowup}
\end{figure}

We also experiment with unfreezing of the LLM backbone, finding that causal effect values blow up beyond their theoretical $[-1, 1]$ bounds in the binary outcome setting. See \autoref{tab:gridsearch_imdb_unfreeze} for grid search details and \autoref{fig:unfreeze_blowup} for an example of the blowup. This issue persisted across unfreezing strategies, including several strategies left out of the gridsearch. 

\subsection*{ATE in the Real IMDB Data}

Though it produced less accurate causal effects on average, we preferred the stabler learning regime of the interleaved experiments for deployment on the observational data. We deploy the interleaved CausalSent model on the real IMDB data, varying regularization loss weights across $0, 0.01, 0.1$. On the real data, we find the effect of the presence of the word "love" on sentiment to be an average $\mathbf{+2.9\%}$ increase in the probability of a positive sentiment label (min $+2.3\%$, max $+3.4\%$). 

\subsection*{ATT Experiments}

We also experiment with the average treatment effect on the treated (ATT). We find that CausalSent struggles to accurately estimate the ATT on synthetic data, especially when the desired "true ATT" is positive. However, as noted in \autoref{sec:att-synth-breakdown}, our synthetic dataset generation is not appropriate for the ATT setting. As a result, we do not trust our ATT results. See \autoref{sec:att-expts} for more details. 

\subsection*{Civil Comments Experiments}

We repeat some of the above experiments against the Civil Comments data, without re-optimizing the model. We find that the model is able to reproduce the \textit{relative} magnitude and direction of causal effects, but overestimates the magnitude of effects by approximately $1.5$-$3$x, which is an unacceptable amount of error. Re-optimization of CausalSent is necessary before deployment on Civil Comments data. It is likely the loss weights and learning rate were inappropriate for the larger dataset and shorter text lengths.

\section{Discussion}

\subsection{Improved Causal Effect Estimates}
\label{sec:improved-effects}

Overall, we were able to achieve significantly better treatment effect prediction accuracy on semi-synthetic IMDB data as compared with \cite{bansal2023rieszcontrolling}'s results on synthetic Civil Comments data. Variations of linear CausalSent with simultaneous learning yield $9.7\%$ to $22.4\%$ MAE relative to target effects on semi-synthetic IMDB data (\autoref{tab:results-1}). In comparison, \cite{bansal2023rieszcontrolling}'s model yields $31.2\%$ to $59.1\%$ MAE relative to target effects on their low overlap ($1\%$) synthetic Civil Comments data\footnote{We call our data semi-synthetic as we do not modify the underlying distribution of confounding texts to induce artificial overlap, and we call \cite{bansal2023rieszcontrolling}'s data synthetic as they artifically manipulate both the treatment effect and the overlap by modeling author intent and using generative sampling to generate new data with the desired effect and overlap. We note that there is no overlap in IMDB's confounding texts prior to feature extraction by BERT, though there is possibly low overlap after feature extraction. However, the extracted feature overlap is not measurable as treatment and confounders cannot be dimensionally separated in the embedded space. Furthermore, overlap calculations would depend on a heuristic tolerance for embedding dissimilarity, since no embedding is perfectly identical. Given the naturally low (to no) overlap in the IMDB data, we compare against \cite{bansal2023rieszcontrolling}'s $1\%$ overlap synthetic Civil Comments setting. }.

\subsection{Case Study of "love"}

\begin{table*}[!htb]
    \centering
    \begin{tabular}{|p{0.8\textwidth}|c|}
        \hline
        \rowcolor{gray!20} \textbf{Random Treated IMDB Movie Review Snippets} & \textbf{Label} \\
        \hline
        \dots evident will to nurture him into their \textcolor{blue}{beloved} practise and hopefully become a good role \dots & 1 \\
        \hline
        \dots image more dramatically. A great success in \textcolor{red}{Love} Actually and as Lizzie in Jane Austen's \dots & 1 \\
        \hline
        \dots meets a beautiful girl and falls in \textcolor{red}{love} so things get even more complicated for \dots & 1 \\
        \hline
        \dots film, and it's one of his most \textcolor{rgb:red,0.1;green,0.8;blue,0.1}{beloved} of all time. Initially a box office \dots ... climactic showdown "the world cannot live without \textcolor{red}{love}" as opposed to the original "you can't \dots & 1 \\
        \hline
        \dots that somewhere in LA is his reincarnated \textcolor{red}{lover} and gives him a junk piece of \dots \newline
        \dots The one thing I \textcolor{blue}{love} about this filthy prostitute Dwars is its \dots & 0 \\
        \hline
        \dots Having seen the trailer, and being a thriller-\textcolor{rgb:red,0.1;green,0.8;blue,0.1}{lover}, I expected to see first of all \dots & 0 \\
        \hline
        \dots Let me just say I \textcolor{rgb:red,0.1;green,0.8;blue,0.1}{loved} the original Boogeyman. Sure, it's a flawed \dots & 0 \\
        \hline
        \dots stars on screen or even if you \textcolor{rgb:red,0.1;green,0.8;blue,0.1}{love} to watch sequels, even if they are \dots & 0 \\
        \hline
    \end{tabular}
    \caption{Random Treated IMDB Examples with Highlighted Treatment Phrase \textcolor{red}{\textbf{love}}. We highlight positive attachments of "love" to the movie (or concepts related to the movie) in \textcolor{rgb:red,0.1;green,0.8;blue,0.1}{green}, neutral usages in \textcolor{red}{red}, and positive attachments of "love" to unrelated objects in \textcolor{blue}{blue}.}
    \label{tab:love-examples}
\end{table*}

As noted in \autoref{sec:sentiment-analysis}, the presence of the word "love" is correlated with an $+18.4\%$ increase in positive sentiment labels across the IMDB data. Our causal effect estimate of $+2.9\%$ is much lower. Does this seem realistic? 

Let us take a closer look at a random selection of treated IMDB examples in \autoref{tab:love-examples}, with the treatment phrase \textcolor{red}{\textbf{love}} highlighted. We assume that "love" only yields a positive effect on the movie review sentiment if it is (1) used in a positive context and (2) the object of the word "love" is the movie, or directly related to the movie. We highlight positive attachments of "love" to the movie in \textcolor{rgb:red,0.1;green,0.8;blue,0.1}{green}, neutral usages in \textcolor{red}{red}, and positive attachments of "love" to unrelated objects in \textcolor{blue}{blue}. Notice that across the positive examples, most of the occurrences of "love" are either neutral or not directed at the movie being reviewed. At the same time, "love" attaches positive meaning to the movie or related concepts in several of the negative reviews, (e.g. "thiller-lover") but these positive effects are insufficient to determine the overall review labels. Other random samples of the treated data yielded similar findings: the word "love" is used in a wide variety of contexts, and direct links between "love" and positive sentiment are rare.  

While the above is merely a qualitative analysis, we feel that the estimated ATE of $+2.9\%$ is realistic, whereas the naive correlation estimate of $+18.4\%$ is likely partially spurious.

\subsection{Model Sensitivity}
\label{sec:sensitivity}

It took numerous iterations of architecture modification and grid search experiments to derive a model which produced good causal effect estimates for the IMDB data. Furthermore, applying this same model setup off-the-shelf to Civil Comments yielded poor causal effect estimates. This suggests that model architecture and hyperparameter optimization are sensitive and data specific.

\subsection{Suggestions for Textual Effect Modeling}
\label{sec:suggestions}

Despite model sensitivity, we feel that the ATE estimation performance of CausalSent on IMDB text data is highly promising. Thus, CausalSent serves as a solid start for researchers hoping to estimate causal effects in text data. We recommend starting with our simultaneous or interleaved learning settings, using linear Riesz and sentiment heads, freezing a single shared LLM encoder backbone, tuning learning rate and loss weights appropriately for your data, and validating the model against synthetic examples prior to applying it observationally. To improve estimates, retrain models several times with varying loss weights or varied random seeds, and ensemble model predictions. 

To apply a similar model to estimate the ATT, an appropriate synthetic data generation algorithm needs to be derived for the purposes of validation experiments (see \autoref{sec:att-synth-breakdown}).

\subsection{Future Directions}
\label{sec:future-directions}

Numerous avenues remain to improve causal effect estimation in text classification. Riesz estimation overfitting bias should be mitigated by implementing cross-fitting as in \cite{chernozhukov2021riesznet}. More complex Riesz representer heads can be learned alongside a more complex aversarial loss minimization, as proposed in \cite{chernozhukov2020adversarial}. Other treatments should be considered (words vs. antonyms, words vs. same part of speech neutrals, sentence/phrase treatments, etc.). \cite{bansal2023rieszcontrolling}'s \texttt{FEAG} algorithm to produce and train against synthetic data should be re-explored on IMDB using improved feature effect estimates from CausalSent. As evidenced by \autoref{tab:love-examples}, treatment effects are highly heterogenous and depend on context. As such, extensions of the present architecture to estimate heterogoneous treatment effects would likely be more informative and performant. Additional observational experiments should be performed with the IMDB CausalSent model to understand the average effect of various phrases on review sentiment, with possible economic implications for businesses hoping to better understand their own product reviews. Lastly, an appropriate synthetic data generation algorithm should be devised for the ATT, and the ATT setting should be further explored.

\subsection{Limitations}
\label{sec:limitations}

The chief limitation of the present work is that one of our critical assumptions (\autoref{sec:assumptions}), overlap, is violated. The high dimensionality and sparsity of text (exacerbated by the length of IMDB movie reviews) makes overlap very small or zero \cite{damour2020overlap}. However, with the right model settings, we achieve good effect estimates regardless.

Additionally, our definition of treatment is problematic. Prepending words to text yields artificial examples which would not naturally occur. Extensions to support ATT estimation would largely mitigate this issue, since synthetic counterfactual inputs could be generated via in-place phrase masking/ replacement. This would be more appropriate with respect to language constructs as opposed to arbitrary treatments such as prepending, post-pending, or LLM-inserting treatment words into the data. 

%\begin{enumerate}
%    \item \textbf{Overlap is questionable}: It is hard to understand/believe overlap in the black-box embedding space.
%    \item \textbf{Causal graph is questionable}: The generative process for language is complex and not well-understood.
%    \item \textbf{Treatment choice and heuristic selection}: Selection of the treatment word is critical, yet somewhat arbitrary and hard to believe.
%    \item \textbf{High-dimensional challenges}: It was extremely difficult to design an architecture which learned accurate causal effects in this high-dimensional text space, especially while trying to simultaneously learn the outcome (sentiment). Counterfactual dataset augmentation techniques might be a better approach. Effect estimates are nonetheless noisy to learn and don\'t achieve super clean convergence.
%    \item \textbf{Noisy regularization}: Our implementation of a running ATE calculation was necessary for simultaneous learning but could be smoothed and improved to yield a less noisy ATE graph throughout training. Regularization appears periodically noisy due to small amounts of data included in the ATE calculation, making it noisy in non-interleaved training regimes.
%\end{enumerate}

\section{Conclusion}

Estimating accurate causal effects in high-dimensional text data is extremely challenging, but critical to developing interpretable NLP models. We focus on extending the work of \cite{bansal2023rieszcontrolling} to learn a RieszNet-style text classification model with superior feature effect estimation performance, achieving significantly lower MAE on our semi-synthetic IMDB data compared to the prior model's MAE on synthetic Civil Comments data (\autoref{sec:improved-effects}). 

ATE estimation is highly sensitive to model architecture and hyperparameters (\autoref{sec:sensitivity}). As such we recommend using our work as a starting point, validating ATE estimation on synthetic data, and subsequently ensembling several validated models on real data (\autoref{sec:suggestions}). 

Though we are able to achieve good causal effect estimation performance on IMDB movie review data, we propose numerous future directions which are likely to vastly improve the accuracy and consistency of the current model (\autoref{sec:future-directions}). 

\section{Code}

All code for the present work can be found at \url{https://github.com/danielfrees/causalsent}.

\clearpage

\newpage
\bibliography{cs328}
\bibliographystyle{icml2024}

\clearpage

\appendix

\section{Assumptions}
\label{sec:assumptions}

For our causal effect estimations to be valid, we need three assumptions to hold. 

\begin{definition}[Ignorability]
The treatment assignment \( T \) is said to be ignorable given covariates \( X \) if 
\[
Y(0), Y(1) \perp T \mid X,
\]
where \( Y(0) \) and \( Y(1) \) are the potential outcomes.
\end{definition}

\begin{definition}[Overlap]
The overlap assumption requires that for all \( x \in \mathcal{X} \),
\[
0 < P(T = 1 \mid X = x) < 1,
\]
ensuring that each individual has a non-zero probability of being assigned to either treatment group.
\end{definition}

\begin{definition}[Identification]
A causal effect is identifiable if the assumptions of ignorability and overlap hold, allowing the average treatment effect (ATE) to be expressed as:
\[
\text{ATE} = \mathbb{E}[\mathbb{E}[Y \mid X, T = 1] - \mathbb{E}[Y \mid X, T = 0]].
\]
\end{definition}

\section{Synthetic Dataset Generation Details}
\label{sec:synthetic-alg}

\begin{algorithm}[H]
\label{alg:synth-data}
\caption{Create Synthetic Dataset with Target ATE}
\begin{algorithmic}[1]
\Require 
\Statex Text data $\{\textbf{X}, \textbf{Y}\}$, treatment phrase $s$, proportion treated $p_t$, target ATE $\tau$, append position $a$, ignore case $c$
\State $n \gets$ length of $\textbf{X}$
\State Randomly select treated indices $T \subseteq \{1,\dots,n\}$ with $|T| = \text{round}(p_t \cdot n)$
\State Untreated indices $U \gets \{1,\dots,n\} \setminus T$
\For{each $i \in T$}
    \State Add phrase $s$ to $\textbf{X}_i$ at position $a$ (if not already present)
\EndFor
\For{each $i \in U$}
    \State Remove phrase $s$ from $\textbf{X}_i$ (if present)
\EndFor
\State Compute initial ATE: $\hat{\tau} \gets \mathbb{E}[Y | T=1] - \mathbb{E}[Y | T=0]$
\State $\Delta \tau \gets \tau - \hat{\tau}$
\If{$\Delta \tau > 0$}
    \State Flip labels from $0$ to $1$ in treated group and/or from $1$ to $0$ in untreated group to increase ATE by $\Delta \tau$
\ElsIf{$\Delta \tau < 0$}
    \State Flip labels from $1$ to $0$ in treated group and/or from $0$ to $1$ in untreated group to decrease ATE by $|\Delta \tau|$
\EndIf
\State Update labels $\textbf{Y}$ accordingly
\State \Return Modified dataset $\{\textbf{X}, \textbf{Y}\}$
\end{algorithmic}
\end{algorithm}

To understand why this process achieves our desired synthetic ATE, consider the following rough proof.

\begin{proof}
\label{pf:label-flipping}
We aim to create a synthetic dataset where the Average Treatment Effect (ATE) matches a specified target $\tau$ by flipping binary outcome labels in treated and untreated groups.

\textbf{Definitions:}

Let $T$ be the treatment indicator where $T=1$ for treated instances and $T=0$ for untreated instances. Let $\textbf{Y}$ represent the binary outcome labels. The Average Treatment Effect (ATE) is defined as:
\[
\tau = \mathbb{E}[Y | T=1] - \mathbb{E}[Y | T=0].
\]

\textbf{Initial ATE:}

After randomly assigning treatments to a proportion $p_t$ of the data, we have that $T \perp\!\!\!\perp X$ and thus under our causal graph, the initial ATE can be computed as:
\[
\hat{\tau} = \mathbb{E}[Y | T=1] - \mathbb{E}[Y | T=0].
\]
The difference between the target and the initial ATE is given by:
\[
\Delta \tau = \tau - \hat{\tau}.
\]

\textbf{Effect of Label Flipping:}

Flipping a label in the treated group changes $\mathbb{E}[Y | T=1]$ by $\pm \frac{1}{n_T}$, where $n_T$ is the number of treated instances. Similarly, flipping a label in the untreated group changes $\mathbb{E}[Y | T=0]$ by $\pm \frac{1}{n_U}$, where $n_U$ is the number of untreated instances.

The net change in the ATE is therefore:
\[
\Delta \hat{\tau} = \frac{k_1}{n_T} - \frac{k_0}{n_U},
\]
where $k_1$ is the number of labels flipped in the treated group and $k_0$ is the number of labels flipped in the untreated group.

\textbf{Target ATE:}

To achieve the target ATE $\tau$, the adjusted ATE must satisfy:
\[
\hat{\tau}_{\text{new}} = \hat{\tau} + \Delta \hat{\tau} = \tau.
\]
Substituting $\Delta \hat{\tau}$, we have:
\[
\hat{\tau} + \frac{k_1}{n_T} - \frac{k_0}{n_U} = \tau.
\]
Rearranging, we obtain:
\[
\frac{k_1}{n_T} - \frac{k_0}{n_U} = \Delta \tau.
\]

By solving for $k_1$ and $k_0$, we can determine the exact number of labels to flip in each group to achieve the target ATE.

The label-flipping process is modeled as a Bernoulli trial, where each label has a fixed probability of being flipped:
- For treated labels, $p = \frac{k_1}{n_T}$.
- For untreated labels, $p = \frac{k_0}{n_U}$.

By the law of large numbers, as the sample size increases, the actual proportion of flipped labels converges to the expected proportion. Consequently, the empirical ATE converges to the target ATE $\tau$. Alternatively, we can simply select $k_1$ and $k_0$ indices of labels to flip randomly, and guarantee the exact ATE. 
\end{proof}

\subsection{Synthetic Data Generation for ATT} 
\label{sec:att-synth-breakdown}

We note that if the chosen treatment word pre-exists in the texts (which will be true of every text in the ATT setting), then the current synthetic pipeline breaks down, because we no longer have that $T \perp\!\!\!\perp X$. Our code might suggest that you have achieved your target ATE but in reality one can no longer measure ATE via the simple direct estimator $\mathbb{E}[Y | T=1] - \mathbb{E}[Y | T=0]$, since $X$'s correlation with $T$ will bias the result significantly. To recover the original setting, we could mask all texts of the treatment word, and return to the completely artificial setting of prepending an unseen-before word to a proportion of texts, and following the label flipping algorithm. However, this cancels the potential benefit of investigating an ATT: with an ATT setting we might better adhere to language conventions because the treated word occurs naturally, rather than being prepended to texts. In that setting, we hope to measure the effect of retaining the word vs. masking it (or replacing with an antonym or neutral word). If future works seek to employ a model framework similar to CausalSent, but focused on the ATT, we suggest a more complex synthetic data generation algorithm which somehow mitigates multicollinearities between $T$ and $X$ while minimally affecting the underlying data distribution.

\section{Additional Causal NLP Examples}
\label{sec:add-causal-nlp-ex}

\begin{figure}[h!]
    \centering
    \includegraphics[width=0.8\linewidth]{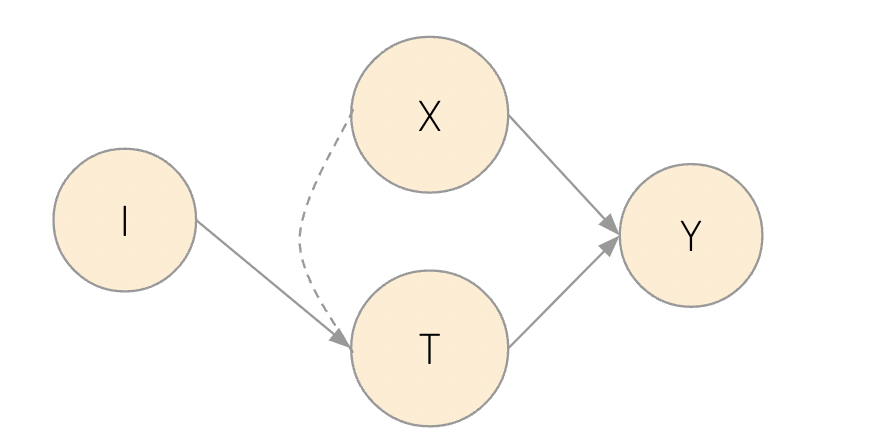}
    \caption{Instrumental Variable model for Movie Review task with browser logs-data}
    \label{fig:instrumental_variable_model}
\end{figure}

Let us consider a couple brief examples to better visualize the task of causal inference in NLP. First, consider the task of understanding treatment effects of specific phrases in online product reviews in \autoref{fig:instrumental_variable_model}. We have logs data $I$, review texts $Z$, and rating outcomes $Y$. We further split the review texts into confounding words $X$ and treatment phrase $T$. We assume that the logs data contains webpage information which is uncorrelated with the ratings, yet might influence the treatment phrase $T$ due to recency bias. Under these assumptions, we can employ instrumental variable techniques to achieve identification. In this example, text serves as both treatment and confounder, as well as the instrumental variable.

\begin{figure}[h!]
    \centering
    \includegraphics[width=0.6\linewidth]{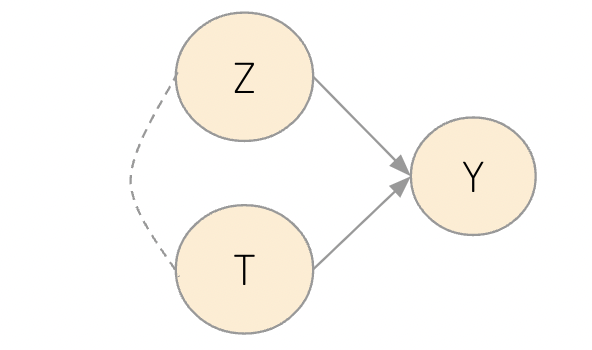}
    \caption{Causal NLP graph with multiple variable modalities}
    \label{fig:multimodal_causal_graph}
\end{figure}

As another example, consider the task of understanding the causal effect of social media profile gender on likes received. Given gender of Twitter profile, $T$, text of tweet, $Z$, likes on the tweet, $Y$, our goal is to measure effect of gender $T$ on likes $Y$ \cite{feder2021causalNLP}. We assume that gender might affect the text within the Tweet. In this case, only the confounder, Tweet content $Z$, is text. To achieve identification we need to condition on $Z$, most likely via propensity-weighting or Riesz Representer based methods.

\section{LLM Transfer Learning}
\label{sec:llm-transfer-learning}

\begin{figure}[h!]
    \centering
    \includegraphics[width=1\linewidth]{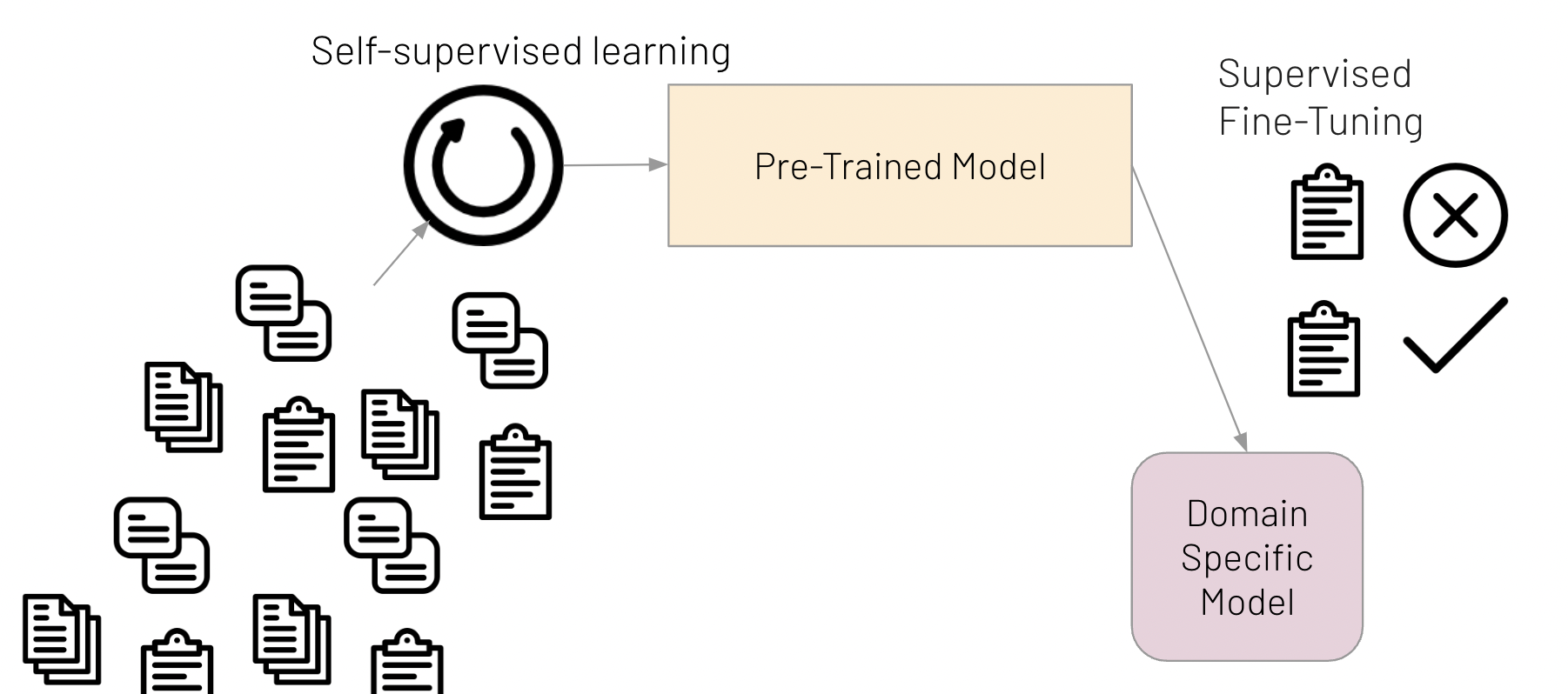}
    \caption{NLP Transfer Learning Overview}
    \label{fig:transfer_learning_overview}
\end{figure}

We utilize a pre-trained LLM as a feature extractor in the present work. 
This can be considered an example of \textit{transfer learning}, a major paradigm within NLP where one trains a massive, transformer-based LLM on a large corpus of generic textual data to learn general language structures and textual understanding, typically through the use of massive datasets and self-supervised learning tasks. This resource-intensive process only needs to be performed once, and then the pre-trained LLM can be applied to many different problems and datasets by fine-tuning the base model further against a much smaller supervised fine-tuning (SFT) dataset, often using parameter-efficient techniques such as transfer layers or low-rank attention mechanism projections (LoRA) \cite{hu2021lora}. This paradigm enables huge efficiency gains in the SFT process, despite generally negligible changes in model performance compared to full fine-tuning \cite{houlsby2019parameter}. A simple transfer learning diagram is illustrated in Figure \ref{fig:transfer_learning_overview}.

\section{Grid Searches}
\label{sec:all-grid-searches}

See \autoref{tab:synth_sim_gridsearch_experiment}, \autoref{tab:gridsearch_civcom}, \autoref{tab:gridsearch_imdb_interleaved}, \autoref{tab:gridsearch_imdb_unfreeze}, and \autoref{tab:gridsearch_obs_ate_imdb} for details on our main grid search experiments. 

\begin{table*}[h!]
\centering
\caption{Hyperparameters for Synthetic, Simultaneous Grid Search}
\label{tab:synth_sim_gridsearch_experiment}
\begin{tabular}{|l|l|}
\hline
\textbf{Category} & \textbf{Hyperparameters} \\ \hline
\textbf{Fixed}    & 
\begin{tabular}[c]{@{}l@{}}
Pretrained Model: \texttt{sentence-transformers/msmarco-distilbert-base-v4} \\
Unfreeze Backbone: \texttt{top0} \\
Riesz Head Type: \texttt{linear} \\
Lambda Regularization (\(\lambda_{\text{reg}}\)): 0 \\
Lambda Riesz (\(\lambda_{\text{riesz}}\)): 0.1 \\
Epochs: 10 \\
Limit Data: 0 \\
Max Sequence Length: 150 \\
Learning Rate: \(5 \times 10^{-5}\) \\
Treatment Phrase: \texttt{asparagus} \\
Dataset: \texttt{IMDB} \\
Log Frequency: 100 \\
Adjust ATE: Enabled \\
Synthetic ATE Treat Fraction: 0.5 \\
Doubly Robust: Enabled \\
Running ATE Option: \{\texttt{--running\_ate}\} \\
\end{tabular} \\ \hline
\textbf{Varied}   & 
\begin{tabular}[c]{@{}l@{}}
Lambda BCE (\(\lambda_{\text{BCE}}\)): \{0, 0.1, 1.0\} \\
Sentiment Head Type: \{\texttt{fcn}, \texttt{linear}\} \\
Synthetic ATE: \{-0.50, -0.25, 0.25, 0.50\} \\
\end{tabular} \\ \hline
\end{tabular}
\end{table*}

\begin{table*}[h!]
\centering
\caption{Hyperparameters for Synthetic Simultaneous Learning on Civil Comments Dataset}
\label{tab:gridsearch_civcom}
\begin{tabular}{|l|l|}
\hline
\textbf{Category} & \textbf{Hyperparameters} \\ \hline
\textbf{Fixed}    & 
\begin{tabular}[c]{@{}l@{}}
Pretrained Model: \texttt{sentence-transformers/msmarco-distilbert-base-v4} \\
Unfreeze Backbone: \texttt{top0} \\
Riesz Head Type: \texttt{linear} \\
Sentiment Head Type: \texttt{linear} \\
Lambda Riesz (\(\lambda_{\text{riesz}}\)): 1.0 \\
Epochs: 10 \\
Limit Data: 30,000 examples \\
Max Sequence Length: 150 \\
Learning Rate: \(5 \times 10^{-5}\) \\
Treatment Phrase: \texttt{asparagus} \\
Dataset: \texttt{Civil Comments} \\
Log Frequency: 50 \\
Adjust ATE: Enabled \\
Synthetic ATE Treat Fraction: 0.5 \\
Doubly Robust: Enabled \\
Running ATE Option: \{\texttt{--running\_ate}\} \\
\end{tabular} \\ \hline
\textbf{Varied}   & 
\begin{tabular}[c]{@{}l@{}}
Lambda BCE (\(\lambda_{\text{BCE}}\)): \{0.1, 1.0\} \\
Lambda Regularization (\(\lambda_{\text{reg}}\)): \{0, 0.01\} \\
Synthetic ATE: \{-0.50, -0.25, 0.25, 0.50\} \\
Interleave Option: \{\texttt{--interleave}, empty\} \\
\end{tabular} \\ \hline
\end{tabular}
\end{table*}

\begin{table*}[h!]
\centering
\caption{Hyperparameters for Synthetic Interleaved Training on IMDB Dataset}
\label{tab:gridsearch_imdb_interleaved}
\begin{tabular}{|l|l|}
\hline
\textbf{Category} & \textbf{Hyperparameters} \\ \hline
\textbf{Fixed}    & 
\begin{tabular}[c]{@{}l@{}}
Pretrained Model: \texttt{sentence-transformers/msmarco-distilbert-base-v4} \\
Unfreeze Backbone: \texttt{top0} \\
Riesz Head Type: \texttt{linear} \\
Lambda Riesz (\(\lambda_{\text{riesz}}\)): 1.0 \\
Epochs: 10 \\
Limit Data: 500 examples \\
Max Sequence Length: 150 \\
Learning Rate: \(5 \times 10^{-5}\) \\
Treatment Phrase: \texttt{cranberry} \\
Dataset: \texttt{IMDB} \\
Log Frequency: 100 \\
Adjust ATE: Enabled \\
Synthetic ATE Treat Fraction: 0.5 \\
Doubly Robust: Enabled \\
Interleaved Training: Enabled \\
Running ATE: Enabled \\
\end{tabular} \\ \hline
\textbf{Varied}   & 
\begin{tabular}[c]{@{}l@{}}
Lambda BCE (\(\lambda_{\text{BCE}}\)): \{0.1, 1.0\} \\
Lambda Regularization (\(\lambda_{\text{reg}}\)): \{0, 0.01\} \\
Sentiment Head Type: \{\texttt{fcn}, \texttt{linear}\} \\
Synthetic ATE: \{-0.50, -0.25, 0.25, 0.50\} \\
\end{tabular} \\ \hline
\end{tabular}
\end{table*}

\begin{table*}[h!]
\centering
\caption{Hyperparameters for Synthetic Interleaved Training with Backbone Unfreezing on IMDB Dataset}
\label{tab:gridsearch_imdb_unfreeze}
\begin{tabular}{|l|l|}
\hline
\textbf{Category} & \textbf{Hyperparameters} \\ \hline
\textbf{Fixed}    & 
\begin{tabular}[c]{@{}l@{}}
Pretrained Model: \texttt{sentence-transformers/msmarco-distilbert-base-v4} \\
Unfreeze Backbone: \texttt{top1} \\
Riesz Head Type: \texttt{linear} \\
Sentiment Head Type: \texttt{linear} \\
Lambda BCE (\(\lambda_{\text{BCE}}\)): 1.0 \\
Lambda Riesz (\(\lambda_{\text{riesz}}\)): 1.0 \\
Epochs: 10 \\
Limit Data: 2000 examples \\
Max Sequence Length: 150 \\
Learning Rate: \(5 \times 10^{-5}\) \\
Treatment Phrase: \texttt{iceberg} \\
Dataset: \texttt{IMDB} \\
Log Frequency: 10 \\
Adjust ATE: Enabled \\
Synthetic ATE Treat Fraction: 0.5 \\
Doubly Robust: Enabled \\
Interleaved Training: Enabled \\
Running ATE: Enabled \\
\end{tabular} \\ \hline
\textbf{Varied}   & 
\begin{tabular}[c]{@{}l@{}}
Lambda Regularization (\(\lambda_{\text{reg}}\)): \{0, 0.01\} \\
Synthetic ATE: \{-0.50, -0.25, 0.25, 0.50\} \\
\end{tabular} \\ \hline
\end{tabular}
\end{table*}

\begin{table*}[h!]
\centering
\caption{Hyperparameters for Observational ATE Grid Search on IMDB Dataset}
\label{tab:gridsearch_obs_ate_imdb}
\begin{tabular}{|l|l|}
\hline
\textbf{Category} & \textbf{Hyperparameters} \\ \hline
\textbf{Fixed}    & 
\begin{tabular}[c]{@{}l@{}}
Pretrained Model: \texttt{sentence-transformers/msmarco-distilbert-base-v4} \\
Unfreeze Backbone: \texttt{top0} \\
Riesz Head Type: \texttt{linear} \\
Sentiment Head Type: \texttt{linear} \\
Lambda BCE (\(\lambda_{\text{BCE}}\)): 1.0 \\
Lambda Riesz (\(\lambda_{\text{riesz}}\)): 1.0 \\
Epochs: 30 \\
Limit Data: All examples \\
Max Sequence Length: 400 \\
Learning Rate: \(5 \times 10^{-5}\) \\
Treatment Phrase: \texttt{love} \\
Dataset: \texttt{IMDB} \\
Log Frequency: 50 \\
Doubly Robust: Enabled \\
Interleaved Training: Enabled \\
Running ATE: Enabled \\
\end{tabular} \\ \hline
\textbf{Varied}   & 
\begin{tabular}[c]{@{}l@{}}
Lambda Regularization (\(\lambda_{\text{reg}}\)): \{0, 0.01, 0.1\} \\
\end{tabular} \\ \hline
\end{tabular}
\end{table*}

\section{ATT Experiments}
\label{sec:att-expts}

See \autoref{tab:gridsearch_att_imdb} for our semi-synthetic ATT IMDB experiment details and \autoref{tab:synth_att_results} for the results, as well as \autoref{tab:gridsearch_obs_att_imdb} for the final observational ATT gridsearch, which finds a very small negative effect of the presence of the word "love" on treated examples (AVG: $-0.66\%$ Pr[POSITIVE], MIN = $-0.77\%$, MAX = $-0.59\%$). Note, however, that we do not trust this result given our poor synthetic data setup for the ATT and resulting lack of validation of the model prior to observational deployment.

\begin{table}[H]
\centering
\begin{tabular}{|c|c|}
\hline
\textbf{True ATT\footnotemark} & \textbf{Estimated (Min / Max)} \\
\hline
-0.5 & -0.203 (-0.2035 / -0.2029) \\
-0.25 & -0.186 (-0.186 / -0.185) \\
0.25 & -0.066 (-0.0664 / -0.0660) \\
0.5 & 0.029 (0.029 / 0.029) \\
\hline
\end{tabular}
\caption{Semi-Synthetic ATT with Interleaved Learning}
\label{tab:synth_att_results}
\end{table}
\footnotetext{Not actually the true ATT since our synthetic data algorithm does not support the ATT setting. See \autoref{sec:att-synth-breakdown}.}

\begin{table*}[h!]
\centering
\caption{Hyperparameters for Synthetic Interleaved Training for ATT on IMDB Dataset}
\label{tab:gridsearch_att_imdb}
\begin{tabular}{|l|l|}
\hline
\textbf{Category} & \textbf{Hyperparameters} \\ \hline
\textbf{Fixed}    & 
\begin{tabular}[c]{@{}l@{}}
Pretrained Model: \texttt{sentence-transformers/msmarco-distilbert-base-v4} \\
Unfreeze Backbone: \texttt{top0} \\
Riesz Head Type: \texttt{linear} \\
Sentiment Head Type: \texttt{linear} \\
Lambda BCE (\(\lambda_{\text{BCE}}\)): 1.0 \\
Lambda Riesz (\(\lambda_{\text{riesz}}\)): 1.0 \\
Epochs: 10 \\
Limit Data: All examples \\
Max Sequence Length: 150 \\
Learning Rate: \(5 \times 10^{-5}\) \\
Treatment Phrase: \texttt{love} \\
Dataset: \texttt{IMDB} \\
Log Frequency: 10 \\
Adjust ATE: Enabled \\
Synthetic ATE Treat Fraction: 0.5 \\
Doubly Robust: Enabled \\
Interleaved Training: Enabled \\
Running ATE: Enabled \\
Treated Only: Enabled \\
\end{tabular} \\ \hline
\textbf{Varied}   & 
\begin{tabular}[c]{@{}l@{}}
Lambda Regularization (\(\lambda_{\text{reg}}\)): \{0, 0.01\} \\
Synthetic ATE: \{-0.50, -0.25, 0.25, 0.50\} \\
\end{tabular} \\ \hline
\end{tabular}
\end{table*}

\begin{table*}[h!]
\centering
\caption{Hyperparameters for Observational ATT Grid Search on IMDB Dataset}
\label{tab:gridsearch_obs_att_imdb}
\begin{tabular}{|l|l|}
\hline
\textbf{Category} & \textbf{Hyperparameters} \\ \hline
\textbf{Fixed}    & 
\begin{tabular}[c]{@{}l@{}}
Pretrained Model: \texttt{sentence-transformers/msmarco-distilbert-base-v4} \\
Unfreeze Backbone: \texttt{top0} \\
Riesz Head Type: \texttt{linear} \\
Sentiment Head Type: \texttt{linear} \\
Lambda BCE (\(\lambda_{\text{BCE}}\)): 1.0 \\
Lambda Riesz (\(\lambda_{\text{riesz}}\)): 1.0 \\
Epochs: 30 \\
Limit Data: All examples \\
Max Sequence Length: 400 \\
Learning Rate: \(5 \times 10^{-5}\) \\
Treatment Phrase: \texttt{love} \\
Dataset: \texttt{IMDB} \\
Log Frequency: 50 \\
Doubly Robust: Enabled \\
Interleaved Training: Enabled \\
Running ATE: Enabled \\
Treated Only: Enabled \\
\end{tabular} \\ \hline
\textbf{Varied}   & 
\begin{tabular}[c]{@{}l@{}}
Lambda Regularization (\(\lambda_{\text{reg}}\)): \{0, 0.01, 0.1\} \\
\end{tabular} \\ \hline
\end{tabular}
\end{table*}

\clearpage

\clearpage

\section{Regularizing Fine-Tuned Embeddings for Text Matching}
\label{ref:sec:regularize-embed}

We initially explored regularization of fine-tuned embedding models for the task of recommending paper citations, using a dataset we generated for ACL abstract text matching. The goal was to determine whether interventional regularization techniques like \cite{bansal2022usinginterventionsimproveoutofdistribution} might be effective in the long-form text setting. However, the model was too computationally complex to train on our available hardware. 

\subsection{Literature Review}

\subsubsection{Intervention-based Methods}

 \cite{bansal2022usinginterventionsimproveoutofdistribution} investigate an intervention-based method for using causal effects to constrain fine-tuned models to be similar to their base encoder models. They find that fine-tuned encoders perform worse than base models on out-of-distribution (OOD) data such as new item categories or queries. To combat this, their interventional strategy consists of masking words in texts and producing the text embeddings before and after masking. The cosine similarity of these two embeddings is then calculated, with a high similarity implying that the word has a small effect on the embedding. The results of Bansal et al show that causal-based regularization can reduce the deleterious effects of fine-tuning, and the authors recommend their method for mild to moderate shifts in data distributions.

\subsubsection{Algorithmic and Heuristic Methods}

Similarly, \cite{pang2021match} aim to minimize spurious noise in text inputs towards developing a superior semantic encoder for text matching. In contrast to \cite{bansal2022usinginterventionsimproveoutofdistribution}, their approach is algorithmic, using PageRank to filter noisy chunks of words in the attention blocks of a Transformer-based encoder model. Each pair of sentences is chopped into chunks, and those chunks are fed into a graph where each node is a text chunk and each edge is weighted based on the semantic similarity between chunks. Low scoring chunks are considered heuristically to be likely spurious content, as suggested by \cite{pang2017deeprank}. The work of \cite{pang2021match} further differs from both \cite{bansal2023rieszcontrolling} and \cite{bansal2022usinginterventionsimproveoutofdistribution} in that it focuses on long texts, which is a generally harder problem both pedagogically and computationally. Pang et al.'s results demonstrate that isolating semantically relevant inputs can improve not only out-of-distribution performance, but overall text-matching performance. 

Whereas both \cite{bansal2023rieszcontrolling} and \cite{bansal2022usinginterventionsimproveoutofdistribution} regularize language models during fine-tuning by adding a loss term proportional to the difference in causal effects between the pre-trained and fine-tuned model, \cite{pang2021match} uses an approach that filters out unimportant chunks of text for each sentence pair.  This represents a fundamental difference in the regularization approach and suggests that both data filtering and loss function modifications can be useful causal regularization strategies. However, it seems likely that data filtering might be more effective for tasks involving long texts. Further, based on the results of \cite{bansal2023rieszcontrolling}, data filtering approaches such as this are likely to harm performance on certain examples where the extra data might have provided useful signal for the downstream prediction task. 

Most importantly, \cite{pang2021match}'s results suggest that causal (or heuristically causal) regularization methods are likely to be effective in the long-form text matching domain as well. Given the current research interest in mining and modeling of huge texts, this result is very promising. A better understanding of causal regularization methods might prove hugely impactful not only for short-form text maching and sentiment analysis, but also for larger scale LLM training and large-scale text data mining. 

\subsection{Architecture and Training}

The below architecture was used to train a fine-tuned DistilBERT, and regularize it against a frozen DistilBERT (since pre-trained models are more invariant to OOD shifts, but lag in performance for specific text matching tasks). We trained via SimCSE loss and planned to also optionally employ interventional feature effect regularization by computing feature effects as absolute decreases in self-similarity when dropping out a particular text feature from one side of the self-similarity dot product. 

\begin{figure}[H]
    \centering
 \includegraphics[width=0.9\linewidth]{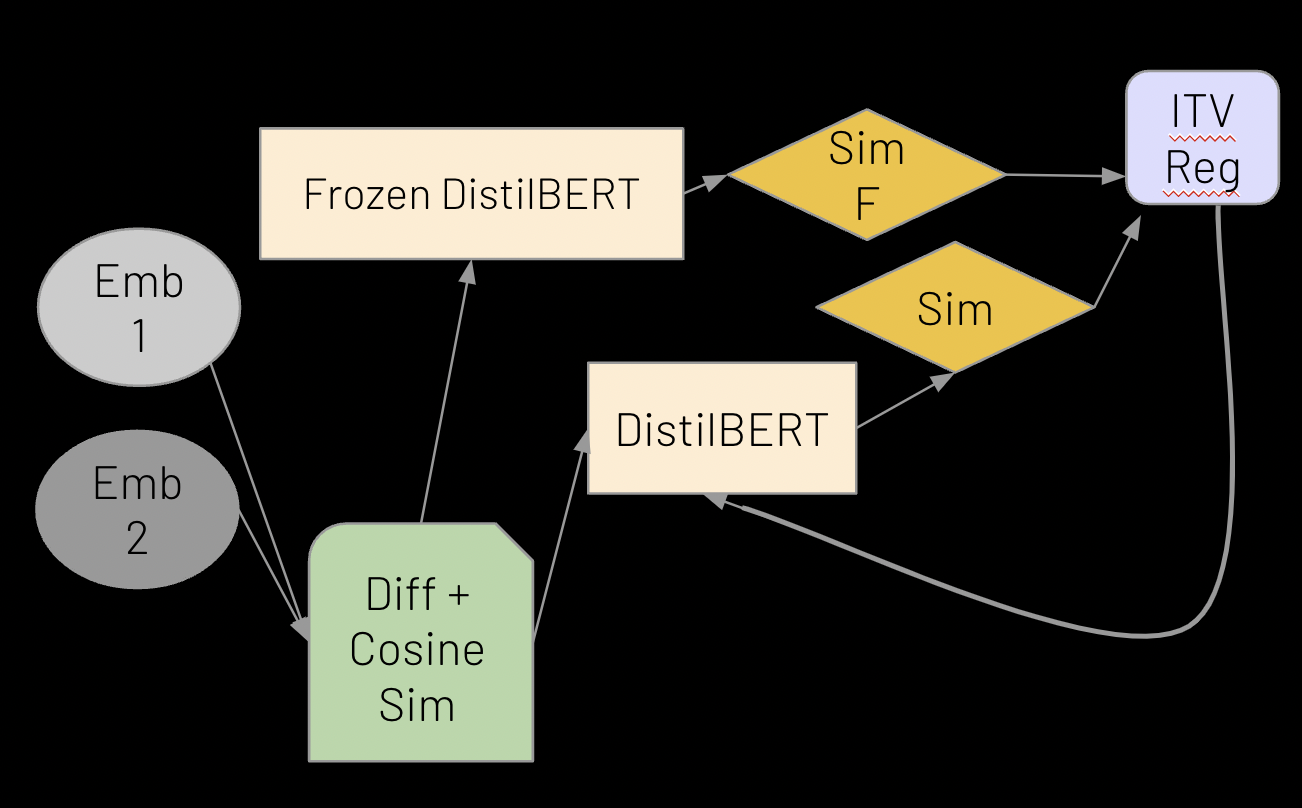}
    \caption{SimDistilBERT Model Architecture}
\end{figure}

\clearpage

\end{document}